\documentclass{article}

    \PassOptionsToPackage{numbers, compress}{natbib}
\usepackage{microtype}
 \usepackage[preprint]{neurips_2026}

\usepackage[utf8]{inputenc} 
\usepackage[T1]{fontenc}    
\usepackage{hyperref}       
\usepackage{url}            
\usepackage{booktabs}       
\usepackage{amsfonts}       
\usepackage{nicefrac}       
\usepackage{microtype}      
\usepackage{xcolor}         

\usepackage{array}
\usepackage{makecell}
\usepackage{enumitem}
\usepackage{multirow} 
\usepackage{overpic}
\usepackage{amsmath}
\usepackage{orcidlink}
\usepackage{booktabs}   
\usepackage{tabularx}   
\usepackage{siunitx}    
\usepackage{makecell}   
\hypersetup{
    colorlinks=false,
    pdfborder={0 0 0}
}

\title{ThinkJEPA: Empowering Latent World Models with Large Vision--Language Reasoning Models}

\author{%
\textbf{Haichao Zhang}\textsuperscript{1} \quad
\textbf{Yijiang Li}\textsuperscript{2} \quad
\textbf{Shwai He}\textsuperscript{3} \quad
\textbf{Tushar Nagarajan}\textsuperscript{4} \\
\textbf{Mingfei Chen}\textsuperscript{5} \quad
\textbf{Jianglin Lu}\textsuperscript{1} \quad
\textbf{Ang Li}\textsuperscript{3} \quad
\textbf{Yun Fu}\textsuperscript{1} \\
\\[-0.25em]
{\small
\textsuperscript{1}Northeastern University \quad
\textsuperscript{2}University of California San Diego \quad
\textsuperscript{3}University of Maryland
} \\
{\small
\textsuperscript{4}The University of Texas at Austin \quad
\textsuperscript{5}University of Washington
} \\
\\[-0.5em]
{\footnotesize
\texttt{\{zhang.haich,lu.jiang\}@northeastern.edu} \quad
\texttt{yunfu@ece.neu.edu}
} \\
{\footnotesize
\texttt{yijiangli@ucsd.edu} \quad
\texttt{\{shwaihe,angliece\}@umd.edu}
} \\
{\footnotesize
\texttt{tushar.nagarajan@utexas.edu} \quad
\texttt{lasiafly@uw.edu}
}
}

\begin{document}

\maketitle

\vspace{-2.0em}
\begin{center}
{\small
Project Page: \url{https://www.zhanghaichao.xyz/ThinkJEPA/} \\
Code: \url{https://github.com/Hai-chao-Zhang/ThinkJEPA}
}
\end{center}

\begin{abstract}
Recent progress in latent world models (e.g., V-JEPA2) has shown promising capability in forecasting future world states from video observations. Nevertheless, dense prediction from a short observation window limits temporal context and can bias predictors toward local, low-level extrapolation, making it difficult to capture long-horizon semantics and reducing downstream utility. Vision--language models (VLMs), in contrast, provide strong semantic grounding and general knowledge by reasoning over uniformly sampled observation frames, but they are not ideal as standalone dense predictors due to compute-driven sparse sampling, a language-output bottleneck that compresses fine-grained interaction states into text-oriented representations, and a data-regime mismatch when adapting to small action-conditioned datasets.
We propose a VLM-guided JEPA-style latent world modeling framework that combines dense-frame dynamics modeling with long-horizon semantic guidance via a dual-temporal pathway: a dense JEPA branch for fine-grained motion and interaction cues, and a uniformly sampled observation VLM \emph{thinker} branch with a larger temporal stride for knowledge-rich guidance. To transfer the VLM's progressive reasoning signals effectively, we introduce a hierarchical pyramid representation extraction module that aggregates multi-layer VLM representations into guidance features compatible with latent prediction. 
Across EgoDex, EgoExo4D, BAIR Robot Pushing, and Physion, ThinkJEPA outperforms diverse latent world model and trajectory prediction baselines across egocentric trajectory prediction, long-horizon rollout, robotic latent prediction, and physical-scene forecasting.
These results show that broad visual--semantic guidance from a VLM thinker can benefit JEPA-style latent forecasting.
\end{abstract}

\section{Introduction}
\label{sec:intro}

World models~\cite{zhang2026SurveyPhysicalAI} aim to learn predictive abstractions of the environment that support forecasting, planning, and control.
Among them, \emph{latent} world models are particularly appealing: by predicting in representation space, they avoid generating photorealistic pixels or detailed 3D geometry, which can be computationally expensive and often unnecessary for downstream decision making.
This paradigm, exemplified by JEPA-style methods (e.g., V-JEPA2~\cite{assran2025v}), promises improved efficiency and encourages the model to emphasize higher-level structure (e.g., dynamics and physical constraints) rather than overfitting to appearance.

Despite strong progress in V-JEPA2~\cite{assran2025v} and its variants, existing JEPA-style latent world models still face two key limitations.
\textbf{(1) Limited temporal perspective for prediction.}
Most approaches rely on a short observation window consisting of densely sampled observation frames to predict future latents.
While dense sampling captures fine-grained motion, it restricts temporal context and can bias the predictor toward local dynamics, missing longer-horizon semantics and event-level cues that are critical for robust forecasting.
\textbf{(2) Weak semantic grounding and general knowledge alignment.}
The latent space is typically learned via self-supervised visual representation learning (often related to masked reconstruction/prediction objectives), which yields motion-sensitive features but provides limited alignment to open-vocabulary concepts and compositional knowledge.
As a result, the predictor may model \emph{how} things move without understanding \emph{what} the entities are and \emph{which attributes or relations} matter, limiting generalization beyond a narrow domain (e.g., a single manipulation dataset).

A natural alternative is to leverage modern vision-language models (VLMs), which excel at high-level video understanding~\cite{tang2025video,Feng_2025_ICCV} and reasoning due to large-scale pretraining and multimodal alignment.
When applied to uniformly sampled observation frames with a larger temporal stride, VLMs can capture long-range context, recognize entities and their attributes, and draw upon general world knowledge~\cite{zhang2025linkedout} that is often missing from purely visual latent predictors.
This complementary capability motivates a promising direction: \emph{using a VLM as a thinker to guide latent world modeling.}
However, directly using VLMs as standalone dense predictors is often impractical and can be suboptimal in representation for fine-grained dynamics.
\textbf{Compute-driven sparsity.}
Video VLMs operate under quadratic attention cost and GPU memory constraints, and thus typically process only a small number of uniformly sampled observation frames.
This design provides long-horizon context but makes it difficult to model high-FPS, fine-grained dynamics crucial for physical interaction and manipulation.
\textbf{Language-output bottleneck.}~\cite{pikabea2025breaking}
Most VLM pipelines ultimately produce \emph{language} outputs (e.g., captions, rationales, or action descriptions).
To generate text, visual information is progressively transformed through stacked transformer layers toward language-generation objectives and discrete token prediction.
This induces an output bottleneck: fine-grained spatial details and continuous interaction states (e.g., contact, precise trajectories, fast motions) are compressed into a language-compatible representation, which is effective for semantic recognition but often inadequate for accurate physical forecasting.
Consequently, language-based planning with VLM outputs can be coherent in text yet physically inconsistent.
\textbf{Data regime mismatch.}~\cite{xiao2025videoqa}
Moreover, deploying VLMs for domain-specific prediction or control often requires adaptation to relatively small, domain-specific datasets, where naïve fine-tuning can hurt general knowledge and semantic capabilities (e.g., catastrophic forgetting~\cite{zhai2023investigating}).

These observations suggest that VLMs are best used as \emph{semantic and knowledge-guidance providers}, rather than standalone dense predictors.
We therefore propose to \emph{integrate a VLM-thinker branch into a JEPA-style latent world model}, combining dense-frame dynamics modeling with long-horizon semantic guidance in a unified framework.
Specifically, we retain the dense-frame observation pathway of V-JEPA-style models to preserve fine-grained motion and interaction cues, while introducing a second branch that feeds uniformly sampled observation frames to a VLM to obtain long-horizon, knowledge-rich guidance.
These VLM signals are injected into the JEPA predictor to improve semantic grounding and enhance the generalization of future latent prediction.

A further challenge is \emph{how} to extract useful guidance from a VLM.
Using only the final-layer VLM features is often suboptimal: deeper layers are increasingly shaped toward language-generation objectives, while intermediate layers can contain richer visual reasoning signals with better spatial sensitivity.
Motivated by this observation, we introduce a \textbf{hierarchical pyramid representation extraction} module that aggregates multi-depth VLM representations and distills them into guidance features compatible with the JEPA predictor, enabling the predictor to benefit from the VLM's progressive reasoning process rather than a single terminal representation.

Our contributions are summarized as follows:
\begin{itemize}[nosep,leftmargin=*]
    \item We propose a \textbf{VLM-guided JEPA-style latent world model} that integrates a VLM as a \emph{thinker} to provide semantic grounding and general knowledge guidance for future latent prediction.
    \item We design a \textbf{dual-temporal pathway} for observation frames: (i) a dense-frame JEPA pathway for fine-grained dynamics modeling, and (ii) a uniformly sampled VLM pathway with a larger temporal stride to capture long-horizon context and high-level concepts.
    \item We introduce a \textbf{hierarchical pyramid representation extraction} module that aggregates multi-layer VLM features to preserve visual reasoning cues and inject them into the JEPA predictor.
    \item Extensive experiments across EgoDex, EgoExo4D, BAIR Robot Pushing, and Physion show that ThinkJEPA consistently improves over diverse latent world model and trajectory prediction baselines, demonstrating strong performance across egocentric trajectory prediction, long-horizon rollout, robotic latent prediction, and physical-scene forecasting.
\end{itemize}
\section{Related Works}
\label{sec:related}

\textbf{Latent World Models.}
Latent world models \cite{ha2018world,hafner2020dreamcontrollearningbehaviors,hafner2020dreamerv2} aim to learn predictive abstractions of the environment that support forecasting, planning, and control. 
By modeling dynamics in a learned representation space, these approaches enable efficient prediction of future states without explicitly generating high-dimensional observations. 
Early works explore latent dynamics learning for video prediction \cite{ebert2017visual} and model-based reinforcement learning \cite{Hafner2023MasteringDD}, where a compact latent state captures the temporal evolution of observations. 
Recent advances in predictive representation learning further strengthen this paradigm. 
In particular, JEPA-style approaches \cite{LeCun2022APT,assran2023selfsupervisedlearningimagesjointembedding} learn representations through predictive objectives that encourage models to capture higher-level structure such as motion patterns and physical interactions. 
Recent systems such as V-JEPA2 demonstrate the scalability of this approach and show promising results for video understanding and world modeling tasks.
Despite these advances, most latent world models are learned solely from visual signals and lack alignment with open-vocabulary semantics or external knowledge, which can limit their ability to incorporate higher-level cues for complex forecasting scenarios.

\textbf{Vision-Language Models.}
Vision-language models (VLMs) have achieved remarkable progress in multimodal representation learning by aligning visual and textual modalities using large-scale image–text data \cite{Radford2021LearningTV,Li2022BLIPBL,Li2023BLIP2BL,zhang2024pixels,zhang2025unified}. 
Early approaches focus on joint representation learning and multimodal understanding tasks such as image captioning and visual question answering. 
More recent multimodal large language models (MLLMs) extend pretrained language models to process visual tokens, enabling instruction following and multimodal reasoning capabilities \cite{alayrac2022flamingo,huang2023language,li2024core}. 
Representative systems such as LLaVA series \cite{Liu2023VisualIT,Li2024LLaVAOneVisionEV} integrate vision encoders with large language models through projection layers or cross-attention mechanisms. 
While these models demonstrate strong semantic reasoning and multimodal understanding capabilities, they are primarily designed for perception and reasoning tasks, and are not optimized for modeling structured physical dynamics. 

\textbf{Multimodal Fusion.} Language has increasingly been used as a high-level control signal for visual generation and decision-making systems. 
Text-conditioned generative models enable natural language prompts to guide image synthesis and editing, as demonstrated by diffusion-based approaches such as DALL·E, Imagen, and Diffusion Transformers (DiT) \cite{Ramesh2022HierarchicalTI,10.5555/3600270.3602913,Peebles2022ScalableDM}. 
Language guidance has also been explored in embodied decision-making frameworks, where large language models provide high-level instructions or goals for perception and action \cite{saycan2022arxiv}. 
These works highlight the potential of language as a flexible interface for controlling visual and embodied systems. 
However, leveraging language signals to guide structured physical forecasting remains relatively underexplored. 
\noindent\textbf{JEPA-style predictors with VLMs.}
Recent work has explored combining language models with JEPA-style representations, but largely in directions that differ from latent world modeling.
For example, VL-JEPA~\cite{chen2025vl} incorporates language signals into a joint-embedding predictive framework, and other approaches use V-JEPA representations as inputs to large language models for video understanding~\cite{assran2025v}.
While effective for multimodal understanding, these designs often shift the primary output interface toward language generation or do not explicitly maintain a latent forecasting interface for downstream world-model tasks.
In contrast, ThinkJEPA retains JEPA-style latent forecasting and leverages VLM semantics as \emph{guidance} by injecting VLM-derived features into the JEPA predictor, preserving dense latent prediction while adding long-horizon semantic cues.

\begin{figure*}[t]
    \centering
    \includegraphics[width=\textwidth]{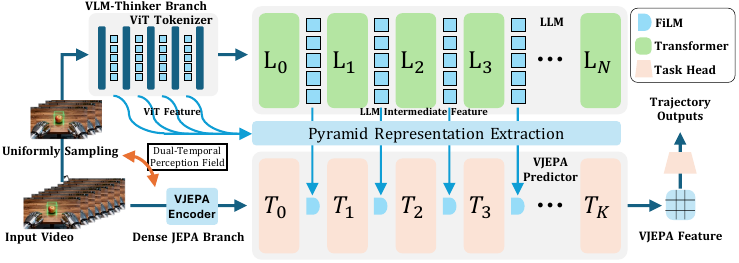}
    \caption{\small \textbf{Overall Architecture of ThinkJEPA.}
    ThinkJEPA couples a dense JEPA branch for fine-grained latent dynamics modeling with a uniformly sampled VLM-\emph{thinker} branch that provides long-horizon semantic guidance.
    The VLM guidance---including visual tokens from the ViT visual tokenizer and intermediate hidden states from the language model---is distilled by a \emph{pyramidal representation extraction} module and injected into the V-JEPA predictor via layer-wise modulation.
    Concretely, guidance derived from language-model layers $\{L_0,\dots,L_N\}$ is mapped to modulation parameters for predictor layers $\{T_0,\dots,T_K\}$.
    The predicted future latents are concatenated with past teacher latents to form the full latent sequence, which is then fed into a task head to produce downstream trajectory predictions.}
    \label{fig:thinkjepa_overall}
\end{figure*}

\section{Methodology}
\label{sec:method}

\subsection{Problem Definition}
\textbf{Basic Settings.}
Given a video clip $v$ with $N$ frames, our goal is to forecast future latent representations that support downstream tasks; in this work, we focus on 3D hand trajectory prediction.
We adopt a JEPA-style latent world modeling paradigm: a visual backbone encodes video frames into latent tokens, and a transformer predictor forecasts future latent tokens from past observations.
To improve semantic grounding and long-horizon reasoning, we further condition the predictor on cached features from a video VLM \emph{thinking} model (we use \texttt{Qwen3-VL (Thinking)} in our implementation), which serves as a \emph{thinker} providing knowledge-rich guidance.

\textbf{Long-Horizon Latent Forecasting via Recursion.}
\label{sec:recursive_forecast}
For long videos where the forecasting horizon exceeds the clip length supported by a single forward pass, we adopt the standard recursive rollout strategy commonly used in JEPA-style predictors.
Concretely, the predictor takes the latent tokens forecast in the previous step as input for the next step, enabling iterative rollout of future latents beyond the original window.
Although recursion allows arbitrarily long-horizon forecasting, it is susceptible to error accumulation over time.
Accordingly, we evaluate both one-shot forecasting and recursive rollouts in our experiments, and analyze robustness under long-horizon prediction.

\subsection{Dual-Temporal Perception Field Sampling Architecture}
\label{sec:dual_temporal}
A central challenge in combining VLM reasoning with latent world modeling is the mismatch between (i) the \emph{dense temporal signal} required for accurate dynamics forecasting and (ii) the \emph{long-horizon temporal context} required for semantic understanding and event-level reasoning.
Dense sampling preserves high-frequency motion and interaction cues but typically covers only a short time span, whereas sparse uniform sampling covers a long time span but discards dense motion details.
To reconcile this trade-off under practical compute and memory budgets, ThinkJEPA adopts a \textit{dual-temporal perception-field} design that explicitly assigns these two roles to two complementary branches.

Given an input video clip $v=\{I_t\}_{t=1}^{N}$ with $N$ frames as \textit{observation}, we construct two temporally sampled inputs:
(i) a uniformly sampled clip $v_u$ for the VLM-\emph{thinker} branch, providing a large temporal perception field for global context and semantics; and
(ii) a densely sampled clip $v_d$ for the JEPA branch, providing high-frequency temporal cues for fine-grained latent forecasting.
The two branches are synchronized at the sample level (derived from the same $v$) and later fused through layer-wise guidance injection (Sec.~\ref{sec:method}).

\textbf{Large temporal perception field sampling for the VLM thinker branch.}
Video VLMs are powerful for semantic grounding because they can identify entities, attributes, and event-level relationships by leveraging large-scale multimodal pretraining.
However, applying transformer-based VLMs to long videos is constrained by quadratic attention cost and GPU memory usage, which typically limits the number of frames that can be processed in a single forward pass.
As a result, VLMs commonly adopt \emph{uniform temporal sampling}: a small set of frames is selected to span a long time horizon.
Although this choice inevitably discards dense motion details, it maximizes temporal coverage and enables the VLM to reason over long-range context.
In ThinkJEPA, we follow this practice and use the VLM branch specifically for long-horizon semantics and knowledge guidance (rather than dense dynamics prediction).
We use \texttt{Qwen3-VL (Thinking)} as the VLM thinker and cache its intermediate representations for efficient conditioning of the latent predictor.
Formally, we define the uniformly sampled clip
\begin{equation}
v_u \;=\; \{ I_{s_i} \}_{i=1}^{N_u}, 
\qquad
s_i \;=\; \left\lfloor 1 + (i-1)\cdot \frac{N-1}{N_u-1} \right\rfloor,
\label{eq:uniform_sampling}
\end{equation}
where $N_u$ is the number of sampled frames for the VLM thinker branch.
This sampling spans the entire clip, providing a large temporal perception field under limited compute.

\textbf{Dense frame sampling for the JEPA branch.}
In contrast, JEPA-style latent world modeling requires dense temporal observations to accurately forecast future latents.
Fine-grained dynamics, contact changes, and subtle interactions are often expressed as high-frequency temporal signals that are poorly captured by sparse sampling.
Therefore, ThinkJEPA uses a dense sampling strategy for the JEPA branch and restricts it to a shorter observation window, where all frames are retained.
Formally, we define an observation window starting at frame index $t_0$ and construct the dense clip
\begin{equation}
v_d \;=\; \{ I_t \}_{t=t_0}^{t_0+N_d-1},
\label{eq:dense_sampling}
\end{equation}
where $N_d$ is the number of densely sampled frames.
The V-JEPA backbone encodes $v_d$ into per-frame patch tokens, producing past latent tokens $F^{\text{past}}$.
A JEPA-style predictor then forecasts future latent tokens $\hat{F}^{\text{fut}}$ from $F^{\text{past}}$.
These predicted latents serve as the target representation for downstream heads (e.g., trajectory regression), while the VLM branch provides complementary long-horizon semantic guidance to improve grounding and generalization.

\textbf{Why dual-temporal sampling matters.}
The uniform VLM sampling and dense JEPA sampling are not redundant: they target different failure modes.
Uniform sampling enables the VLM thinker to access long-range context and semantics that are difficult to infer from a short dense window, whereas dense sampling enables accurate modeling of high-frequency dynamics that sparse VLM inputs cannot represent reliably.
By coupling these two perception fields and injecting VLM guidance into the JEPA predictor, ThinkJEPA benefits from both long-horizon semantic context and fine-grained dynamic cues in future latent forecasting.

\subsection{JEPA-style latent tokenization and forecasting}
\label{sec:jepa}
The visual backbone encodes a densely sampled clip into per-frame spatial tokens
$F \in \mathbb{R}^{B \times T \times P \times D}$, where $B$ is the batch size, $T$ is the number of frames in the observation window, $P$ is the number of spatial tokens per frame, and $D$ is the backbone latent dimension.
We split the clip into past and future segments and use a masked-token transformer predictor to forecast future latent tokens from past tokens.
The predictor operates in an internal dimension $D_p$ and projects its outputs back to the backbone latent space of dimension $D$.

\textbf{Rollout of the JEPA branch.}
\label{sec:jepa_rollout}
Densely sampled inputs provide strong motion and interaction cues, but they also limit the temporal duration that can be processed in a single forward pass due to compute and memory constraints.
For videos whose length exceeds the JEPA observation window, we therefore perform \emph{recursive rollout} by repeatedly forecasting the next segment and feeding the predicted latents into the subsequent step.

Let $W$ denote the number of frames per JEPA window (e.g., $W=T_p+T_f$), and let $k$ index rollout steps.
At step $k$, the predictor takes past latent tokens $F^{\text{past}}_{k}$ and outputs future latent tokens $\hat{F}^{\text{fut}}_{k}$:
\begin{equation}
\hat{F}^{\text{fut}}_{k} \;=\; g\!\left(F^{\text{past}}_{k}\right),
\label{eq:jepa_pred_step}
\end{equation}
where $g(\cdot)$ is the JEPA-style predictor.
For the next step, we set the past tokens to be the previously predicted future tokens (or a shifted window that includes them):
\begin{equation}
F^{\text{past}}_{k+1} \;\leftarrow\; \hat{F}^{\text{fut}}_{k}.
\label{eq:jepa_rollout}
\end{equation}
By iterating Eqs.~\eqref{eq:jepa_pred_step}, we can roll out arbitrarily long-horizon latent forecasts.

While rollout enables long-horizon prediction, it is susceptible to \emph{error accumulation} and remains limited by the local temporal context within each window.
This motivates incorporating VLM-thinker guidance, which provides complementary long-horizon semantic context to stabilize forecasting and improve generalization (Sec.~\ref{sec:dual_temporal}).

\subsection{VLM Thinker: Hierarchical Pyramid Representation Extraction}
\label{sec:vlm_thinker}

\textbf{Complementarity via injecting VLM guidance into JEPA.}
\label{sec:vlm_into_jepa}
Prior work has explored combining language and JEPA-style representations in different directions.
For example, VL-JEPA~\cite{chen2025vl} and approaches that feed V-JEPA features into LLMs for video understanding~\cite{assran2025v} primarily treat JEPA features as inputs to a language model.
While effective for video-to-text understanding, this design shifts the output space toward language generation and does not directly preserve a latent world model interface for downstream prediction.
In contrast, our goal is to retain JEPA-style \emph{latent forecasting} while leveraging VLM semantics as \emph{guidance}.
This is non-trivial because the VLM must provide useful long-horizon semantic context without replacing the dense dynamics modeling of the JEPA predictor.

As discussed in Sec.~\ref{sec:dual_temporal}, uniform sampling enables the VLM thinker to access long-range context and event-level semantics under limited compute, whereas dense sampling provides the JEPA branch with high-frequency temporal signals for fine-grained dynamics.
We combine these two pathways by injecting VLM guidance into the JEPA predictor in a layer-wise manner.
Concretely, given a uniformly sampled clip $v_u$ and a densely sampled clip $v_d$, the predictor forecasts future latent tokens conditioned on both VLM guidance and an optional text prompt:
{\small
\begin{equation}
\hat{F}^{\text{fut}} \;=\; g\!\left(F^{\text{past}}(v_d)\,;\, \phi(v_u),\, p\right),
\label{eq:vlm_guided_jepa}
\end{equation}
}
where $F^{\text{past}}(v_d)$ are past latent tokens extracted by the V-JEPA backbone from the dense clip, $\phi(v_u)$ denotes VLM-derived guidance features from the uniform clip, $p$ denotes the text prompt provided to the VLM thinker, and $g(\cdot;\cdot)$ is the V-JEPA predictor.
In practice, the VLM thinker prompt $p$ is generated from a general summarization request, which helps the thinker focus on relevant entities and events.

\textbf{Hierarchical pyramid representation extraction.}
\label{sec:pyramid}
A key question is \emph{which} VLM representations are most suitable for guiding latent forecasting.
Using only the final-layer VLM features can be suboptimal, since deeper layers are increasingly shaped toward language-generation objectives, while intermediate layers often retain richer visual reasoning cues and better spatial sensitivity.
This observation is supported by prior analyses showing that aggregating intermediate LLM representations can outperform using a single terminal layer for downstream tasks (e.g.,~\cite{zhang2025linkedout}).
Moreover, visual tokenizer outputs may lose fine-grained cues after passing through multimodal fusion and language decoding stages.

Motivated by these findings, we propose a \textit{hierarchical pyramid representation extraction} module that aggregates multi-depth VLM signals.
Specifically, we combine (i) visual tokens from the VLM visual encoder (ViT tokenizer) and (ii) intermediate hidden states from selected language-model layers, forming a depth-wise pyramid over the VLM.
These multi-depth features are pooled and projected into the predictor space, yielding guidance features $\phi(v_u)$ that preserve both low-level visual cues and high-level semantic reasoning traces.

\textbf{Layer-wise guidance.}
\label{sec:film_inject}
We inject the extracted thinker guidance into the JEPA predictor via feature-wise linear modulation (FiLM)~\cite{perez2018film}.
For predictor block $\ell$, the guidance produces modulation parameters $(\gamma_\ell,\beta_\ell)$, and we modulate the block input as
{\small\begin{equation}
 \mathrm{FiLM}(z;\gamma_\ell,\beta_\ell) \;=\; \gamma_\ell \odot z + \beta_\ell.
\label{eq:film}
\end{equation}}
This yields layer-wise, sample-specific conditioning that injects semantic and knowledge cues into latent forecasting without requiring the VLM to act as a dense predictor.

\textbf{Joint prediction for downstream regression.}
\label{sec:joint_pred}
For the basic setting, we follow standard V-JEPA downstream protocols~\cite{assran2025v} by feeding the predicted latent tokens into a task head for trajectory regression.
For long-horizon forecasting with recursive rollout (Sec.~\ref{sec:jepa_rollout}), we concatenate the past latents and the predicted future latents into a full-length token sequence, which is then fed to the temporal regression head to produce the target trajectories.

\section{Experiments}
\label{sec:exp}

\subsection{Experimental Setup}
\label{sec:exp_setup}

\textbf{Benchmarks.}
We evaluate ThinkJEPA on four video forecasting benchmarks covering complementary regimes: egocentric trajectory forecasting (EgoDex~\cite{egodex} and EgoExo4D~\cite{grauman2024ego}), robotic latent rollout (BAIR Robot Pushing~\cite{finn2016unsupervised}), and action-free physical-scene forecasting (Physion~\cite{bear1physion}).
EgoDex and EgoExo4D evaluate 3D human/hand trajectory prediction, BAIR evaluates action-conditioned rollout in frozen V-JEPA latent space, and Physion evaluates object-contact prediction from observed and predicted future latents.
Detailed dataset protocols, splits, preprocessing, metric definitions, and implementation details are provided in Sec.~\ref{sec:supp_exp_details}.

\textbf{Observation-only VLM access.}
To avoid future-frame leakage, the VLM branch is restricted to observed frames in all forecasting experiments.
For EgoDex, EgoExo4D, and Physion, uniformly sampled VLM frames are drawn only from the observation window; for BAIR, the VLM observes only the initial block $x_0{:}x_4$, which is reused for all rollout steps.
No future labels or target-frame metadata are provided to the VLM branch for BAIR or Physion.

\textbf{Baselines.}
We compare ThinkJEPA with Qwen3-VL~\cite{bai2025qwen3} Thinking (VLM-only), V-JEPA Predictor~\cite{assran2025v} (JEPA-only), DINO-WM~\cite{zhou2025dino}, persistence/copy-last for latent rollout, and task-specific EgoDex trajectory baselines including decoder-only and encoder-decoder predictors trained with Behavior Cloning~\cite{nasiriany2024robocasa}, DDPM~\cite{ho2020denoising}, and Flow Matching~\cite{lipman2022flow}.
Additional baseline details and training settings are provided in Sec.~\ref{sec:supp_baselines}.

\subsection{Experiment Results}

\textbf{Quantitative Comparison:}
\label{sec:exp_main}
Tab.~\ref{tab:main_comparison} reports main comparison on EgoDex and EgoExo4D datasets. On these trajectory forecasting benchmarks, ThinkJEPA consistently outperforms both single-branch baselines in trajectory prediction, achieving substantially lower ADE/FDE and markedly higher Acc.
Compared to the \textbf{V-JEPA Predictor}, injecting VLM-thinker guidance improves semantic grounding while preserving dense dynamics cues, leading to a large gain in downstream trajectory accuracy.
Compared to \textbf{Qwen3-VL Thinking}, ThinkJEPA avoids relying on sparse, language-oriented representations as a standalone predictor and instead uses the VLM as guidance, yielding a more physically grounded forecast.
In addition to trajectory metrics, ThinkJEPA also improves latent forecasting quality (lower FD/SL1/CD), indicating that guidance injection benefits representation prediction rather than only the downstream head.
Overall, these results show that \textbf{ThinkJEPA can surpass both a strong VLM baseline and a strong latent world model baseline} by integrating long-horizon VLM reasoning with dense JEPA-style latent forecasting.
\begin{table}[tb]
\centering
\footnotesize
\setlength{\tabcolsep}{3.0pt}
\renewcommand{\arraystretch}{1.10}
\caption{\small \textbf{Quantitative comparison across datasets.} We report trajectory metrics (ADE/FDE/Acc) and latent forecasting metrics (FD/SL1/CD). FD/SL1/CD denote V-JEPA feature distance, latent SmoothL1, and latent cosine distance, respectively. All values are reported with three decimal places.}
\begin{tabular*}{\linewidth}{@{\extracolsep{\fill}}llcccccc@{}}
\toprule
\textbf{Dataset} & \textbf{Model} &
\textbf{ADE}$\downarrow$ & \textbf{FDE}$\downarrow$ & \textbf{Acc}$\uparrow$ &
\textbf{FD}$\downarrow$ & \textbf{SL1}$\downarrow$ & \textbf{CD}$\downarrow$ \\
\midrule
\multirow{3}{*}{\textbf{EgoDex}} 
& Qwen3-VL Thinking & 0.142 & 0.144 & 0.084 & 99.538 & 1.656 & 0.615 \\
& V-JEPA Predictor  & 0.071 & 0.066 & 0.471 & 74.223 & 1.252 & 0.317 \\
& ThinkJEPA   & \textbf{0.061} & \textbf{0.056} & \textbf{0.596} & \textbf{74.032} & \textbf{1.248} & \textbf{0.315} \\

\midrule
\multirow{3}{*}{\textbf{EgoExo4D}} 
& Qwen3-VL Thinking & 0.661 & 0.690 & 0.038 & 104.548 & 1.756 & 0.690 \\
& V-JEPA Predictor  & 0.659 & 0.636 & 0.074 & 89.244  & 1.520 & 0.469 \\
& ThinkJEPA         & \textbf{0.622} & \textbf{0.597} & \textbf{0.171} & \textbf{79.654} & \textbf{1.364} & \textbf{0.359} \\
\bottomrule
\end{tabular*}
\label{tab:main_comparison}
\end{table}

\textbf{Task-specific trajectory Comparison:}
To compare against direct trajectory prediction methods, we include the EgoDex benchmark baselines~\cite{jia2025x}, which combine decoder-only or encoder-decoder predictors with Behavior Cloning (BC), DDPM, and Flow Matching (FM) objectives. Detailed baseline descriptions are provided in Sec.~\ref{sec:supp_egodex_traj_baselines}.
As shown in Tab.~\ref{tab:egodex_baselines_ade_fde}, ThinkJEPA achieves the best ADE/FDE among all compared methods.
Compared with the strongest task-specific baseline, Decoder-only + BC, ThinkJEPA reduces ADE from 0.0767 to 0.0610 and FDE from 0.0818 to 0.0560.
These results show that VLM-guided latent forecasting provides a stronger trajectory representation than directly predicting trajectories with conventional trajectory prediction heads.

\begin{table}[t]
\centering
\footnotesize
\setlength{\tabcolsep}{4.2pt}
\renewcommand{\arraystretch}{1.08}
\caption{\small \textbf{Comparison with EgoDex trajectory prediction baselines on EgoDex.} We compare ThinkJEPA against the trajectory prediction baselines reported in EgoDex, including decoder-only and encoder-decoder architectures with Behavior Cloning, DDPM, and Flow Matching. ThinkJEPA achieves the best ADE/FDE among all compared methods.}
\begin{tabular*}{\linewidth}{@{\extracolsep{\fill}}llcc@{}}
\toprule
\textbf{Group} & \textbf{Model} & \textbf{ADE}$\downarrow$ & \textbf{FDE}$\downarrow$ \\
\midrule
\multirow{6}{*}{\textbf{Trajectory Baselines}}
& Decoder-only + Behavior Cloning & 0.0767 & 0.0818 \\
& Decoder-only + DDPM             & 0.1148 & 0.1238 \\
& Decoder-only + Flow Matching    & 0.1527 & 0.1574 \\
& Encoder-decoder + Behavior Cloning & 0.0774 & 0.0924 \\
& Encoder-decoder + DDPM             & 0.1272 & 0.1245 \\
& Encoder-decoder + Flow Matching    & 0.1736 & 0.1557 \\
\midrule
\multirow{3}{*}{\textbf{Latent Forecasting}}
& Qwen3-VL Thinking & 0.1420 & 0.1440 \\
& V-JEPA Predictor  & 0.0710 & 0.0660 \\
& ThinkJEPA         & \textbf{0.0610} & \textbf{0.0560} \\
\bottomrule
\end{tabular*}

\label{tab:egodex_baselines_ade_fde}
\end{table}

\noindent\textbf{BAIR Robot Pushing: Action-Conditioned Latent Rollout:}
\label{sec:bair_results}
\textbf{Setup.}
We evaluate action-conditioned latent rollout on BAIR Robot Pushing~\cite{finn2016unsupervised}, where the model observes the initial 5-frame block and autoregressively predicts the next three future latent blocks in frozen V-JEPA space; detailed protocol are provided in Sec.~\ref{sec:supp_bair_protocol}.
\textbf{Results and analysis.}
Tab.~\ref{tab:bair_latent_rollout} reports BAIR latent rollout results.
ThinkJEPA improves over the V-JEPA Predictor on all latent metrics, reducing FD from 68.711 to 67.049, SL1 from 1.117 to 1.086, and CD from 0.301 to 0.285.
It also outperforms DINO-WM and the persistence baseline, supporting the effectiveness of ThinkJEPA for action-conditioned latent rollout in robotic interaction.

\begin{table*}[t]
\centering
\footnotesize
\setlength{\tabcolsep}{3.2pt}
\renewcommand{\arraystretch}{1.08}

\begin{minipage}[t]{0.56\textwidth}
\caption{\textbf{BAIR Robot Pushing latent rollout.} We evaluate three-step autoregressive rollout in frozen V-JEPA latent space using 5-frame blocks. ThinkJEPA uses only the initial observed block $x_0{:}x_4$ for VLM conditioning. }
\centering
\begin{tabular*}{\linewidth}{@{\extracolsep{\fill}}lccc@{}}
\toprule
\textbf{Method} & \textbf{FD/L2}$\downarrow$ & \textbf{SL1}$\downarrow$ & \textbf{CD}$\downarrow$ \\
\midrule
Persistence & 86.385 & 1.511 & 0.430 \\
V-JEPA Predictor & 68.711 & 1.117 & 0.301 \\
DINO-WM & 71.720 & 1.176 & 0.323 \\
ThinkJEPA & \textbf{67.049} & \textbf{1.086} & \textbf{0.285} \\
\bottomrule
\end{tabular*}

\label{tab:bair_latent_rollout}
\end{minipage}
\hfill
\begin{minipage}[t]{0.40\textwidth}
\caption{\textbf{Physion object-contact prediction.} All non-oracle methods use only the observation prefix. ThinkJEPA achieves the best Acc/AUC. }
\label{tab:physion_ocp}
\centering
\begin{tabular*}{\linewidth}{@{\extracolsep{\fill}}lcc@{}}
\toprule
\textbf{Method} & \textbf{Acc}$\uparrow$ & \textbf{AUC}$\uparrow$ \\
\midrule
V-JEPA observed-only  & 0.738  & 0.820 \\
V-JEPA Predictor & 0.741  & 0.821 \\
DINO-WM & 0.648  & 0.705 \\
ThinkJEPA & \textbf{0.754}  & \textbf{0.828} \\
\bottomrule
\end{tabular*}
\end{minipage}
\end{table*}

\begin{table}[bt]
\centering
\footnotesize
\setlength{\tabcolsep}{3.2pt}
\renewcommand{\arraystretch}{1.10}
\caption{\textbf{Ablation studies.} We vary the VLM token sources and the thinker module.
\textbf{Encoder} denotes encoding tokens, \textbf{AR} denotes autoregressive tokens;
\textbf{No-dual} disables the dual-temporal VLM pathway, and \textbf{No-Th} removes the thinker module.
We abbreviate latent metrics as \textbf{FD} (feature distance), \textbf{SL1} (SmoothL1), and \textbf{CD} (cosine distance).}
\begin{tabular*}{\linewidth}{@{\extracolsep{\fill}}l
  S[table-format=1.3]
  S[table-format=1.3]
  S[table-format=1.3]
  S[table-format=3.3]
  S[table-format=1.3]
  S[table-format=1.3]@{}}
\toprule
\textbf{Abl.} &
\multicolumn{1}{c}{\textbf{ADE}$\downarrow$} &
\multicolumn{1}{c}{\textbf{FDE}$\downarrow$} &
\multicolumn{1}{c}{\textbf{Acc}$\uparrow$} &
\multicolumn{1}{c}{\textbf{FD}$\downarrow$} &
\multicolumn{1}{c}{\textbf{SL1}$\downarrow$} &
\multicolumn{1}{c}{\textbf{CD}$\downarrow$} \\
\midrule
Encoder+V-JEPA predictor    & \bfseries 0.128 & \bfseries 0.129 & \bfseries 0.100 & 78.869  & 1.340 & 0.360 \\
Encoder-only  & 0.143 & 0.145 & 0.086 & 102.910 & 1.700 & 0.615 \\
AR+V-JEPA predictor   & 0.128 & 0.130 & 0.098 & \bfseries 78.514 & \bfseries 1.333 & \bfseries 0.356 \\
AR-only & 0.142 & 0.144 & 0.086 & 102.910 & 1.700 & 0.615 \\
No-dual-temporal sampling   & 0.128 & 0.130 & 0.099 & 78.862  & 1.340 & 0.360 \\
No-Th & 0.071 & 0.066 & 0.471 & 74.223 & 1.252 & 0.317 \\
\midrule
ThinkJEPA & \bfseries 0.061 & \bfseries 0.056 & \bfseries 0.596 & \bfseries 74.747 & \bfseries 1.263 & \bfseries 0.324 \\

\bottomrule

\end{tabular*}
\label{tab:ablation_tokens}
\end{table}


\begin{table}[tb]
\centering
\footnotesize
\setlength{\tabcolsep}{3.4pt}
\renewcommand{\arraystretch}{1.10}
\caption{\small  \textbf{VLM layer selection on EgoDex.} We compare guidance derived from different VLM layer selections. \textbf{FD} denotes V-JEPA feature distance, \textbf{SL1} denotes latent SmoothL1, and \textbf{CD} denotes latent cosine distance. }
\begin{tabular*}{\linewidth}{@{\extracolsep{\fill}}l
  S[table-format=1.3]
  S[table-format=1.3]
  S[table-format=1.3]
  S[table-format=3.3]
  S[table-format=1.3]
  S[table-format=1.3]@{}}
\toprule
\textbf{Variant} &
\multicolumn{1}{c}{\textbf{ADE}$\downarrow$} &
\multicolumn{1}{c}{\textbf{FDE}$\downarrow$} &
\multicolumn{1}{c}{\textbf{Acc}$\uparrow$} &
\multicolumn{1}{c}{\textbf{FD}$\downarrow$} &
\multicolumn{1}{c}{\textbf{SL1}$\downarrow$} &
\multicolumn{1}{c}{\textbf{CD}$\downarrow$} \\
\midrule
Last-layer & \bfseries 0.128 & \bfseries 0.130 & \bfseries 0.099 & 78.858 & 1.340 & 0.360 \\
Mid-layer  & 0.128 & 0.131 & 0.098 & \bfseries 78.517 & \bfseries 1.333 & \bfseries 0.356 \\
\midrule
All layers (ThinkJEPA) & \bfseries 0.061 & \bfseries 0.056 & \bfseries 0.596 & \bfseries 74.747 & \bfseries 1.263 & \bfseries 0.324 \\
\bottomrule
\end{tabular*}

\label{tab:layer_selection}
\end{table}

\noindent\textbf{Physion: Action-Free Physical Scene Forecasting:}
\label{sec:physion_results}
\textbf{Setup.}
We evaluate action-free physical-scene forecasting on Physion~\cite{bear1physion}, where models observe only the video prefix and predict future latent states for downstream object-contact prediction; detailed split, and observation protocol are provided in Sec.~\ref{sec:supp_physion_protocol}.
\textbf{Results and analysis.}
Tab.~\ref{tab:physion_ocp} reports Physion object-contact prediction results.
ThinkJEPA achieves the best Accuracy and AUC, improving over the V-JEPA Predictor from 0.741 to 0.754 in Acc and from 0.821 to 0.828 in AUC.
Thus, Physion supports the benefit of ThinkJEPA for action-free physical-scene forecasting.
DINO-WM performs worse under the same split, suggesting that V-JEPA-style latent prediction is a stronger representation backbone.

\textbf{Ablation baselines:}
We further conduct controlled ablations to understand which VLM signals are necessary for effective guidance.
We study two aspects: (i) the contribution of different VLM token sources, and (ii) the effect of VLM layer selection.
Detailed ablation settings are provided in Sec.~\ref{sec:supp_ablation_settings}.

\textbf{Ablation on VLM token sources:}
\label{sec:exp_ablation_tokens}
Tab.~\ref{tab:ablation_tokens} studies the contribution of encoding tokens and autoregressive (AR) tokens.
Single-source VLM guidance is insufficient and can even degrade the JEPA-only predictor, indicating that naively injecting partial VLM signals may introduce incomplete or mismatched conditioning.
Encoding tokens mainly preserve input-side visual-context summaries, whereas AR tokens capture generation-side hidden states.
When either source is used in isolation, it lacks the complementary information needed to guide dense latent forecasting effectively.
In contrast, ThinkJEPA combines both token sources through pyramid representation extraction and layer-wise modulation, enabling the JEPA predictor to receive both visual grounding and generation-side semantic guidance.
This suggests that the gain is not due to adding an arbitrary VLM feature, but arises from the interaction between input-side visual grounding and generation-side VLM hidden states.

\textbf{Ablation on VLM layer selection:}
\label{sec:exp_layer_selection}
Tab.~\ref{tab:layer_selection} compares guidance derived from different VLM layer selections.
We observe a trade-off: last-layer guidance slightly improves trajectory metrics, whereas mid-layer guidance yields better latent forecasting quality.
This is consistent with the intuition that deeper layers are increasingly shaped toward language-generation objectives, while intermediate layers retain richer visual reasoning cues.
The all-layer pyramid variant achieves the strongest overall performance, motivating our hierarchical multi-depth guidance design.

\textbf{Long-horizon recursive rollout.}
\label{sec:exp_rollout_traj}
To evaluate forecasting beyond a single prediction window, we perform recursive rollout and report horizon-specific trajectory errors in Tab.~\ref{tab:rollout_traj}.
Qwen3-VL Thinking degrades sharply at longer horizons, confirming that sparse VLM representations are not reliable standalone dense predictors.
The V-JEPA Predictor remains stable but accumulates error as the rollout horizon increases.
In contrast, ThinkJEPA achieves the best performance across all horizons, suggesting that VLM-guided latent forecasting helps stabilize iterative prediction while preserving dense dynamics modeling.
Detailed rollout settings and additional latent-distance diagnostics are provided in Sec.~\ref{sec:supp_rollout}.

\begin{table*}[t]
\centering
\footnotesize
\setlength{\tabcolsep}{3.0pt}
\renewcommand{\arraystretch}{1.10}
\caption{\small \textbf{Recursive rollout on EgoDex: trajectory error vs.\ horizon.} We perform autoregressive rollout for horizons $H\in\{4,8,16,32\}$. 
A@H and F@H denote ADE@H and FDE@H, respectively; the lower, the better.}
\begin{tabular*}{\textwidth}{@{\extracolsep{\fill}}lcccccccc@{}}
\toprule
\textbf{Model} &
\textbf{A@4} & \textbf{A@8} & \textbf{A@16} & \textbf{A@32} &
\textbf{F@4} & \textbf{F@8} & \textbf{F@16} & \textbf{F@32} \\
\midrule
Qwen3-VL Thinking & 0.140 & 0.819 & 1.375 & 1.026 & 0.143 & 2.850 & 0.286 & 1.092 \\
V-JEPA Predictor  & 0.121 & 0.126 & 0.134 & 0.142 & 0.124 & 0.136 & 0.149 & 0.153 \\
ThinkJEPA         & \textbf{0.071} & \textbf{0.078} & \textbf{0.092} & \textbf{0.111} &
                    \textbf{0.073} & \textbf{0.090} & \textbf{0.118} & \textbf{0.136} \\
\bottomrule
\end{tabular*}
\label{tab:rollout_traj}
\end{table*}
\begin{figure*}[b]
    \centering
    \includegraphics[width=\textwidth]{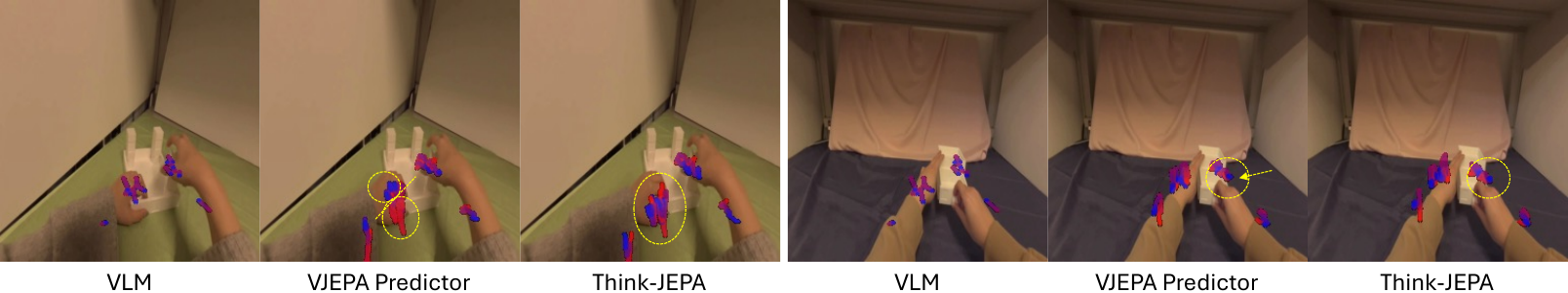}
    \caption{\small \textbf{Qualitative results.} Predicted future hand-manipulation trajectories visualized as heat maps overlaid on the reference frame. Colors indicate temporal progression from blue (earlier) to red (later). Ideally, trajectories transition smoothly from blue to red, indicating coherent motion over time. ThinkJEPA produces smoother trajectories with better temporal consistency and joint alignment.}
    \label{fig:qual}
\end{figure*}

\textbf{Qualitative Results.}
\label{sec:qual}
As shown in Fig.~\ref{fig:qual}, we visualize predicted future hand trajectories by decoding the forecasted V-JEPA latents with the downstream task head and overlaying the resulting 3D joints on a reference frame.
Overall, ThinkJEPA produces more plausible and accurate trajectories: the final endpoint (deep red) aligns more closely with the hand in the reference frame, and the temporal progression is smoother and more diverse over time.
In contrast, as highlighted by the yellow circles, the V-JEPA baseline often exhibits temporally collapsed predictions, where blue points concentrate in a small region, indicating that multiple timesteps and joints are predicted to overlap.
In the first example, the VLM-only baseline hallucinates a non-existent left hand, while the V-JEPA baseline yields less accurate joint localization and noisier motion compared to our method.

\section{Conclusion}
\label{sec:conclusion}
We presented \textbf{ThinkJEPA}, a VLM-guided JEPA-style latent forecasting framework that integrates long-horizon semantic guidance from a vision--language \emph{thinker} with dense latent dynamics prediction.
ThinkJEPA adopts a dual-temporal perception design---uniform sampling over the observed context for the VLM branch and dense sampling for the JEPA branch---and injects pyramid-extracted, multi-depth VLM representations into the JEPA predictor through layer-wise modulation.
This design preserves the latent forecasting interface of JEPA-style world models while enriching future latent prediction with knowledge-aware VLM guidance.
Extensive experiments show that ThinkJEPA consistently outperforms VLM-only, JEPA-only, and task-specific trajectory baselines across four benchmarks: EgoDex, EgoExo4D, BAIR Robot Pushing, and Physion.
These results cover multiple latent world-modeling tasks, including egocentric trajectory prediction, long-horizon recursive rollout, action-conditioned robotic latent prediction, and action-free physical-scene forecasting.
Together, they demonstrate that VLM-guided JEPA-style latent forecasting provides a stronger and more general interface for downstream world-model prediction tasks.
Future work includes extending ThinkJEPA to closed-loop planning, broader embodied tasks, and more scalable VLM-guidance mechanisms for longer and more diverse videos.

%
%
\bibliographystyle{splncs04}
\bibliography{main}

\newpage
\appendix
\section{Additional Experimental Details}
\label{sec:supp_exp_details}

\subsection{Datasets and Protocols}
\label{sec:supp_datasets}

\subsubsection{EgoDex.}
EgoDex~\cite{egodex} is a large-scale benchmark for egocentric dexterous manipulation.
It provides egocentric videos paired with 3D hand and finger pose annotations, making it a natural testbed for latent video forecasting and downstream trajectory regression.
We use the same cached train/test split for all EgoDex methods to ensure fair comparison across VLM-only, V-JEPA-only, ThinkJEPA, and ablation variants.

\subsubsection{EgoExo4D.}
EgoExo4D~\cite{grauman2024ego} is a large-scale multimodal, multiview dataset of skilled human activities.
It contains synchronized egocentric and exocentric videos, with annotations including 3D body pose, 3D hand pose, and gaze.
We use EgoExo4D to evaluate whether ThinkJEPA transfers beyond the EgoDex setting to broader human-motion forecasting from egocentric observations.

\subsubsection{BAIR Robot Pushing.}
BAIR Robot Pushing~\cite{finn2016unsupervised} is used to evaluate action-conditioned latent video rollout in a robotic interaction setting.
We use the \texttt{bair\_robot\_pushing\_small} dataset with $64{\times}64$ videos.
The full TFDS split contains 43,264 training videos and 256 official test videos; in our controlled run, we use 2,000 training videos, 500 validation videos, and the official 256-video test split.
Each sequence uses the first 20 frames and is divided into four 5-frame latent blocks.
The model observes the first block $x_0{:}x_4$ and autoregressively predicts the next three future latent blocks $x_5{:}x_9$, $x_{10}{:}x_{14}$, and $x_{15}{:}x_{19}$ in frozen V-JEPA latent space.
Action-conditioned variants additionally receive robot actions, while visual-only baselines use only observed latent blocks.

\subsubsection{Physion.}
Physion~\cite{bear1physion} is used to evaluate action-free physical scene forecasting.
We use a controlled split with 2,000 training videos, 400 validation videos, and 1,200 test videos across eight scenarios: collide, contain, dominoes, drape, drop, link, roll, and support.
The observation ratio is 0.5: non-oracle methods only observe the prefix of each video and must predict future latent states from the observed context.
Future frames are used only as prediction targets.

\subsubsection{Observation-only VLM access.}
For all datasets, ThinkJEPA restricts VLM access to observed frames only.
For EgoDex, EgoExo4D, and Physion, VLM frames are sampled from the observation window.
For BAIR, the VLM observation is fixed to the initial observed block $x_0{:}x_4$ and reused for all three rollout steps.
Thus, the VLM branch never observes target future frames during forecasting or rollout.
For BAIR and Physion, no future labels or target-frame metadata are provided to the VLM branch.

\subsection{Evaluation Metrics}
\label{sec:supp_metrics}

\subsubsection{Trajectory metrics.}
Let $\hat{Y}\in\mathbb{R}^{B\times T_f \times J \times 3}$ and $Y\in\mathbb{R}^{B\times T_f \times J \times 3}$ denote predicted and ground-truth 3D trajectories over $T_f$ future frames and $J$ joints.
We report ADE, the mean Euclidean distance over all future frames and joints; FDE, the Euclidean error at the final future frame averaged over joints and batch; and threshold Accuracy, the fraction of predicted joint positions whose Euclidean error is below 0.05\,m.

\subsubsection{Latent forecasting metrics.}
For latent prediction, we compare predicted and target V-JEPA latents using FD (feature $\ell_2$ distance), SL1 (SmoothL1 distance), and CD (cosine distance, defined as $1-\cos(\cdot)$).
These metrics measure representation-level forecasting quality independent of the downstream trajectory head.

\subsubsection{Rollout metrics.}
For recursive rollout evaluation, A@H and F@H denote ADE@H and FDE@H at rollout horizon $H\in\{4,8,16,32\}$.
For BAIR latent rollout, we report FD/L2, SL1, and CD over autoregressively predicted latent blocks.

\subsubsection{Physion contact metrics.}
For Physion, we evaluate object-contact prediction (OCP) using binary classification Accuracy and ROC-AUC.
These metrics are distinct from the 0.05\,m trajectory Accuracy used for EgoDex and EgoExo4D.

\subsection{Baselines and Variants}
\label{sec:supp_baselines}

\subsubsection{ThinkJEPA.}
ThinkJEPA uses dense-frame V-JEPA tokens for latent forecasting and injects VLM-thinker guidance derived from both \emph{encoding tokens} and \emph{autoregressive (AR) tokens}.
The VLM guidance is obtained from multi-depth VLM representations and injected into the V-JEPA predictor via layer-wise modulation.

\subsubsection{Qwen3-VL Thinking (VLM-only).}
To isolate the VLM branch, we disable the dense V-JEPA latent input while keeping the VLM branch unchanged.
We then train the same downstream head on VLM-derived representations.
This baseline tests whether long-horizon VLM representations alone can support dense trajectory forecasting under matched task supervision.

\subsubsection{V-JEPA Predictor (JEPA-only).}
We train a V-JEPA predictor and the same downstream head without VLM conditioning.
This baseline represents dense latent forecasting without semantic VLM guidance.

\subsubsection{Persistence / copy-last.}
For latent rollout experiments such as BAIR Robot Pushing, we include a persistence baseline that repeats the last observed latent block for all future rollout steps.
This baseline does not model dynamics and serves as a sanity-check lower bound.

\subsubsection{DINO-WM.}
We compare against DINO-WM~\cite{zhou2025dino}, a latent predictor based on DINOv2 visual latents with a transformer future predictor.
This baseline tests whether the observed gains are specific to the V-JEPA latent space or also hold against another strong self-supervised visual representation.

\subsubsection{Task-specific trajectory baselines.}
For EgoDex trajectory prediction, we additionally compare against decoder-only and encoder-decoder trajectory predictors combined with Behavior Cloning~\cite{nasiriany2024robocasa}, DDPM~\cite{ho2020denoising}, and Flow Matching~\cite{lipman2022flow}, following the EgoDex benchmark protocol.

\subsubsection{Controlled ablations.}
We conduct additional ablations on VLM token sources, VLM layer selections, conditioning mechanisms, and prompt-only VLM inference.
For token-source ablations, \textbf{Encoder} denotes encoding tokens and \textbf{AR} denotes autoregressive tokens.

\section{Model Implementation Details}
\label{sec:impl_details}

\subsection{Backbone.}
We use a V-JEPA-L backbone (\texttt{vit\_large\_rope}) to extract per-frame patch tokens with latent dimension $D{=}1024$.

\subsection{VLM-injected V-JEPA predictor.}
We implement a V-JEPA predictor operating in an internal dimension $D_p{=}384$ and inject VLM-thinker guidance into the predictor via layer-wise FiLM modulation.
We condition each predictor block using $(\gamma_\ell,\beta_\ell)$ derived from cached \texttt{Qwen3-VL (Thinking)} representations.
The cache provides \emph{encoding tokens} and \emph{autoregressive (AR) tokens}, which are projected to $D_p$, pooled, and mapped to per-layer FiLM parameters using lightweight MLP adapters.
For hierarchical pyramid extraction, we cache intermediate hidden states from VLM layers $\mathcal{L}=\{0,4,8,12,16,20,24,27\}$.

\subsection{Trajectory head.}
We use a lightweight temporal trajectory regression head for downstream prediction.
The head first aggregates spatial tokens within each frame via attention pooling with a learnable query, producing a per-frame representation.
It then applies temporal MLP blocks to model cross-frame dependencies, followed by stride-2 temporal downsampling to align the temporal resolution with the prediction horizon.
Finally, a linear projection regresses 3D trajectories with output shape $32 \times 52 \times 3$.

\subsection{Shared hyperparameters.}
Unless otherwise specified, all experiments share a 64-frame input clip at resolution $256\times256$, a past/future split of 32/32 frames, a V-JEPA-L backbone with depth/dim 24/1024, a VLM-injected V-JEPA predictor with dimension $D_p=384$, depth 12, and 6 heads, and a cached \texttt{Qwen3-VL (Thinking)} VLM thinker with token dimension $D_c=2048$.

\subsection{EgoDex Trajectory Prediction Baselines}
\label{sec:supp_egodex_traj_baselines}

Following the EgoDex benchmark protocol~\cite{jia2025x}, we compare against six task-specific trajectory prediction baselines.
These baselines are formed by combining two Transformer architectures with three policy representations.

\subsubsection{Architectures.}
The \textbf{decoder-only} baselines directly predict future trajectories from past observations using a single Transformer-style predictor.
The \textbf{encoder-decoder} baselines first encode the observed context and then decode future trajectories conditioned on the encoded representation.

\subsubsection{Policy representations.}
We evaluate three trajectory modeling objectives:
\textbf{Behavior Cloning (BC)}, which directly regresses future trajectories under supervised imitation;
\textbf{Denoising Diffusion Probabilistic Models (DDPM)}~\cite{ho2020denoising}, which model future trajectory generation through a denoising process;
and \textbf{Flow Matching (FM)}~\cite{lipman2022flow}, which learns a continuous transformation from noise to trajectory samples.
All baselines are trained under the EgoDex 2-second trajectory prediction setting and serve as task-specific references for egocentric hand trajectory forecasting.

\subsection{BAIR Robot Pushing Protocol}
\label{sec:supp_bair_protocol}

We evaluate action-conditioned latent video forecasting on BAIR Robot Pushing~\cite{finn2016unsupervised} using \texttt{bair\_robot\_pushing\_small}.
The full TFDS split contains 43,264 training videos and 256 official test videos; in our controlled run, we use 2,000 training videos, 500 validation videos, and the official 256-video TFDS test split.
Each video is resized to $64{\times}64$, and we use the first 20 frames of each sequence.
We divide each sequence into four 5-frame latent blocks.
The model observes the first block $x_0{:}x_4$ and performs three-step autoregressive rollout to predict $x_5{:}x_9$, $x_{10}{:}x_{14}$, and $x_{15}{:}x_{19}$ in frozen V-JEPA latent space.
Action-conditioned variants additionally receive robot actions, while visual-only baselines use only the observed latent blocks.

\subsection{Physion Protocol}
\label{sec:supp_physion_protocol}

We evaluate action-free physical-scene forecasting on Physion~\cite{bear1physion}.
The benchmark contains eight physical-interaction scenarios: collide, contain, dominoes, drape, drop, link, roll, and support.
We use a controlled split with 2,000 training videos, 400 validation videos, and 1,200 test videos, balanced across scenarios.
The observation ratio is 0.5: non-oracle methods observe only the prefix of each video and must predict future latent states from the observed context.

\subsubsection{Evaluation.}
We evaluate object-contact prediction (OCP) using a binary readout on top of observed tokens and predicted future latent tokens.
We report Accuracy, Balanced Accuracy, and ROC-AUC.
These metrics are distinct from the 0.05\,m trajectory Accuracy used for EgoDex and EgoExo4D.

\subsection{Ablation Settings}
\label{sec:supp_ablation_settings}

\subsubsection{Token-source ablations.}
To assess which VLM token sources contribute to guidance, we evaluate variants that selectively enable different components:
(i) encoding tokens + V-JEPA predictor,
(ii) encoding tokens only,
(iii) AR tokens + V-JEPA predictor,
(iv) AR tokens only,
(v) no dual-temporal VLM pathway,
and (vi) no thinker module.
Here, \textbf{Encoder} denotes encoding tokens extracted from the VLM input-side representations, and \textbf{AR} denotes autoregressive tokens extracted from generation-side hidden states.
The no-thinker variant removes VLM guidance and reduces to the V-JEPA predictor baseline.

\subsubsection{Layer-selection ablations.}
To study the role of hierarchical pyramid extraction, we compare guidance derived from different VLM layer selections, including last-layer guidance, mid-layer guidance, and the full all-layer pyramid guidance used in ThinkJEPA.
All variants follow the same training and evaluation protocol unless otherwise specified.

\subsection{Long-Horizon Rollout Evaluation}
\label{sec:supp_rollout}

To evaluate long-horizon forecasting behavior beyond a single prediction window, we perform recursive rollout.
We use a short-window predictor configuration with $T_p{=}4$ and $T_f{=}4$ for each step and recursively roll out to horizons $H\in\{4,8,16,32\}$.
At each horizon, we report trajectory errors using ADE@H and FDE@H, denoted as A@H and F@H in the main paper.
We also compute Accuracy@H after autoregressive rollout and aggregate latent-distance diagnostics, including L2 distance, SmoothL1 distance, and cosine distance, over the rollout trajectory.

\section{Prompt + Video to VLM-Conditioned Features}
\label{sec:prompt_video_condition}

\subsection{Experimental setting.}
This study evaluates a prompt-conditioned VLM feature design, where the predictor takes cached visual features as the primary input and uses language-modulated VLM features as external conditioning.
The training setup follows the same backbone and downstream trajectory head as the main paper, while changing only the conditioning path.

\subsection{Experimental details.}
The input video is first represented by cached ViT-style spatiotemporal features, which serve as the main predictive substrate.
In parallel, video frames together with a text prompt are passed through \texttt{Qwen3-VL (Thinking)} to extract VLM conditioning features.
The conditioning features include two complementary streams: encoder-side representations and autoregressive generation-side representations.
These VLM features are injected into the predictor rather than replacing the visual backbone.

\subsection{Analysis.}
As shown in Tab.~\ref{tab:prompt_video_condition}, prompt-conditioned VLM features provide a competitive design choice.
Compared with the full ThinkJEPA model, this variant achieves slightly weaker trajectory prediction (ADE/FDE/Acc: 0.069/0.062/0.495 vs.\ 0.061/0.056/0.596), but slightly better latent forecasting metrics (FD/SL1/CD: 74.007/1.248/0.315 vs.\ 74.747/1.263/0.324).
These results suggest that prompt-conditioned VLM features are effective for representation guidance, while the full ThinkJEPA design yields a stronger overall downstream trade-off.

\begin{table}[t]
\centering
\footnotesize
\setlength{\tabcolsep}{3.5pt}
\renewcommand{\arraystretch}{1.08}
\begin{tabular*}{\linewidth}{@{\extracolsep{\fill}}lcccccc@{}}
\toprule
\textbf{Variant} & \textbf{ADE}$\downarrow$ & \textbf{FDE}$\downarrow$ & \textbf{Acc}$\uparrow$ & \textbf{FD}$\downarrow$ & \textbf{SL1}$\downarrow$ & \textbf{CD}$\downarrow$ \\
\midrule
Prompt + video $\rightarrow$ VLM condition & 0.069 & 0.062 & 0.495 & \textbf{74.007} & \textbf{1.248} & \textbf{0.315} \\
ThinkJEPA & \textbf{0.061} & \textbf{0.056} & \textbf{0.596} & 74.747 & 1.263 & 0.324 \\
\bottomrule
\end{tabular*}
\caption{\textbf{Prompt + video to VLM-conditioned features.} The predictor uses cached visual features as the main trajectory backbone and VLM-derived features as external conditioning.}
\label{tab:prompt_video_condition}
\end{table}

\section{Temporal Stride Ablation}
\label{sec:temporal_stride}

\subsection{Experimental setting.}
This study examines the role of temporal sampling granularity in the dual-temporal design.
We compare two temporal strides while keeping the predictor architecture, training budget, and conditioning mechanism fixed.

\subsection{Experimental details.}
EgoDex trajectories are first represented as 64 uniformly sampled temporal points over each episode.
The prediction protocol uses 32 past points and 32 future points.
For \textbf{stride 1}, no temporal decimation is applied, so the predictor observes all 64 sampled points.
For \textbf{stride 2}, the temporal sequence is subsampled before the past/future split, resulting in a coarser temporal representation.
Since the original sequence is already uniformly sampled to 64 points, stride 1 corresponds to denser temporal coverage, while stride 2 reduces temporal resolution.

\subsection{Analysis.}
The results in Tab.~\ref{tab:temporal_stride} show that denser temporal sampling improves both trajectory prediction and latent forecasting quality.
Stride 1 outperforms stride 2 on all reported metrics.
Compared with both stride variants, the full ThinkJEPA model further improves downstream trajectory performance, achieving the best ADE/FDE/Acc overall on this split.

\begin{table}[t]
\centering
\footnotesize
\setlength{\tabcolsep}{3.5pt}
\renewcommand{\arraystretch}{1.08}
\begin{tabular*}{\linewidth}{@{\extracolsep{\fill}}lcccccc@{}}
\toprule
\textbf{Stride} & \textbf{ADE}$\downarrow$ & \textbf{FDE}$\downarrow$ & \textbf{Acc}$\uparrow$ & \textbf{FD}$\downarrow$ & \textbf{SL1}$\downarrow$ & \textbf{CD}$\downarrow$ \\
\midrule
Temporal stride 1 & 0.071 & 0.064 & 0.471 & 73.920 & 1.246 & 0.314 \\
Temporal stride 2 & 0.073 & 0.071 & 0.458 & 74.266 & 1.247 & 0.317 \\
\bottomrule
\end{tabular*}
\caption{\textbf{Temporal stride ablation.} Denser temporal sampling improves trajectory prediction and latent forecasting quality.}
\label{tab:temporal_stride}
\end{table}

\section{Conditioning Mechanism Ablation}
\label{sec:conditioning_ablation}

\subsection{Experimental setting.}
This study compares three conditioning operators under the same backbone, data split, and training budget.
Only the conditioning mechanism is varied.

\subsection{Experimental details.}
We compare three ways of injecting VLM guidance into the predictor:
(i) FiLM-style affine modulation,
(ii) cross-attention conditioning, and
(iii) AdaLN-style adaptive normalization.
All variants consume the same cached VLM features and the same base visual representation stream, so differences can be attributed to the conditioning operator itself.

\subsection{Analysis.}
Tab.~\ref{tab:conditioning_ablation} shows that all three conditioning mechanisms are competitive.
FiLM provides the strongest latent forecasting quality among the three variants, while cross-attention and AdaLN remain close alternatives.
Compared with these controlled variants, the full ThinkJEPA model achieves substantially better trajectory prediction (ADE/FDE/Acc), indicating that the final design used in the paper offers the strongest downstream performance under the current setting.

\begin{table}[t]
\centering
\footnotesize
\setlength{\tabcolsep}{3.5pt}
\renewcommand{\arraystretch}{1.08}
\begin{tabular*}{\linewidth}{@{\extracolsep{\fill}}lcccccc@{}}
\toprule
\textbf{Conditioning} & \textbf{ADE}$\downarrow$ & \textbf{FDE}$\downarrow$ & \textbf{Acc}$\uparrow$ & \textbf{FD}$\downarrow$ & \textbf{SL1}$\downarrow$ & \textbf{CD}$\downarrow$ \\
\midrule
FiLM        & \textbf{0.0706} & \textbf{0.064} & 0.471 & \textbf{73.878} & \textbf{1.245} & \textbf{0.314} \\
Cross-attn  & 0.0707 & 0.066 & \textbf{0.475} & 73.965 & 1.247 & 0.315 \\
AdaLN       & 0.0708 & 0.065 & 0.474 & 74.280 & 1.253 & 0.317 \\
\bottomrule
\end{tabular*}
\caption{\textbf{Conditioning mechanism ablation.} We compare FiLM, cross-attention, and AdaLN under the same training setup.}
\label{tab:conditioning_ablation}
\end{table}

\section{Direct Visual Conditioning and Deepstack-Token Removal}
\label{sec:visual_vs_deepstack}

\subsection{Experimental setting.}
This study compares two variants:
(i) removing the VLM branch entirely and conditioning only on direct visual features, and
(ii) keeping the VLM branch but removing the deepstack/thinking-token contribution.
We further compare both variants against the full ThinkJEPA model.

\subsection{Experimental details.}
For \textbf{direct visual conditioning}, the predictor removes all VLM conditioning and operates only on visual backbone features.
This serves as a controlled visual-only baseline within the same predictor family.
For \textbf{deepstack-token removal}, the VLM branch is preserved, but the generation-side thinking/deepstack token contribution is explicitly dropped before conditioning is consumed by the predictor.
This removal is implemented using token filtering and hard zeroing, ensuring that the removed tokens do not leak through the conditioning path.

\subsection{Analysis.}
The results in Tab.~\ref{tab:visual_vs_deepstack} show that both ablations remain competitive.
Dropping deepstack tokens yields slightly stronger latent forecasting quality than direct visual conditioning alone, suggesting that the full VLM branch contributes non-trivial information.
However, both variants are weaker than the full ThinkJEPA model in downstream trajectory performance, and ThinkJEPA achieves the best ADE/FDE/Acc overall.
This indicates that the complete VLM guidance pathway is most effective when used as part of the full model design.

\subsection{Why FiLM as the default conditioning operator.}
Although we compare multiple conditioning operators in the supplementary experiments, we choose FiLM as the default design in ThinkJEPA because our primary goal is to improve \emph{latent feature prediction}, rather than only optimizing the downstream regression head.
FiLM performs feature-wise modulation directly in the predictor latent space, allowing the VLM thinker to refine the predicted representation while preserving the JEPA-style latent forecasting interface.
Compared with cross-attention, FiLM is lighter-weight and introduces less structural change to the predictor, making it easier to attribute gains to guidance rather than additional token interactions.
Compared with normalization-based conditioning such as AdaLN, FiLM provides a more direct channel-wise control over the latent features themselves, which is particularly aligned with our objective of improving representation-level prediction quality.
For this reason, we adopt FiLM as the main conditioning operator in the paper, while including other variants as complementary ablations.

\begin{table}[t]
\centering
\footnotesize
\setlength{\tabcolsep}{3.5pt}
\renewcommand{\arraystretch}{1.08}
\begin{tabular*}{\linewidth}{@{\extracolsep{\fill}}lccc@{}}
\toprule
\textbf{Variant} & \textbf{ADE}$\downarrow$ & \textbf{FDE}$\downarrow$ & \textbf{Acc}$\uparrow$ \\
\midrule
Direct visual conditioning & 0.071 & 0.066 & 0.475 \\
Drop deepstack tokens      & 0.072 & 0.066 & 0.464 \\
ThinkJEPA                  & \textbf{0.061} & \textbf{0.056} & \textbf{0.596} \\
\bottomrule
\end{tabular*}
\caption{\textbf{Direct visual conditioning vs.\ deepstack-token removal.} Both ablations remain competitive, while ThinkJEPA achieves the strongest downstream trajectory performance.}
\label{tab:visual_vs_deepstack}
\end{table}

\section{Pure Prompt-Only VLM Baseline}
\label{sec:prompt_only}

\subsection{Experimental setting.}
This study evaluates a pure VLM baseline without any task-specific prediction head.
Unlike the VLM-only baseline in the main paper, which uses a trained downstream head on top of VLM-derived features, this study directly prompts the VLM with video and text and asks it to output future 3D trajectories in structured form.
Its purpose is to provide a zero-shot reference point for direct prompting without task-specific adaptation.

\subsection{Experimental details.}
We use \texttt{Qwen3-VL (Thinking)} as a prompt-only baseline.
The model observes only the past segment of the video and is instructed to predict future hand trajectories in world coordinates.
It outputs a small set of future waypoints in JSON format, which are then interpolated to the full prediction horizon for evaluation.
No trajectory head is trained, making this a zero-shot or prompt-only baseline.

\subsection{Analysis.}
As shown in Tab.~\ref{tab:prompt_only}, the pure prompt-only baseline performs dramatically worse than ThinkJEPA, with ADE/FDE of 10.855/10.927 compared to 0.061/0.056 for our method.
This large gap confirms that direct prompting of a general-purpose VLM is insufficient for fine-grained metric-space trajectory prediction.
In addition, parsing success is poor in this setting, indicating that structured trajectory generation itself is unstable under pure prompting.
We therefore regard this baseline as an intentionally weak but informative reference point, rather than a competitive predictor for this benchmark.

\subsection{Implication for the main-paper VLM-only baseline.}
The result in Tab.~\ref{tab:prompt_only} also clarifies why the \emph{VLM-only} baseline in the main paper is implemented with a trained task head rather than direct prompting.
A general-purpose VLM that has not been fine-tuned for future trajectory prediction performs very poorly in this setting, even though it possesses strong general semantic reasoning ability.
This indicates that zero-shot prompting alone is insufficient for fine-grained metric-space forecasting of hand motion.
Therefore, the main-paper \emph{VLM-only} baseline is intentionally designed as a fairer and stronger comparison: it uses the same task-specific training protocol and downstream prediction head, while removing the JEPA latent forecasting pathway.
In this way, the comparison in the main paper isolates the benefit of JEPA-style latent prediction versus VLM-based features under matched supervision and optimization, rather than comparing against an intentionally weak zero-shot prompt baseline.

\begin{table}[t]
\centering
\footnotesize
\setlength{\tabcolsep}{4.0pt}
\renewcommand{\arraystretch}{1.08}
\begin{tabular*}{\linewidth}{@{\extracolsep{\fill}}lccc@{}}
\toprule
\textbf{Baseline} & \textbf{ADE}$\downarrow$ & \textbf{FDE}$\downarrow$ & \textbf{Acc}$\uparrow$ \\
\midrule
Qwen3-VL prompt-only & 10.855 & 10.927 & 0.000 \\
ThinkJEPA            & \textbf{0.061} & \textbf{0.056} & \textbf{0.596} \\
\bottomrule
\end{tabular*}
\caption{\textbf{Pure prompt-only VLM baseline.} We directly prompt \texttt{Qwen3-VL (Thinking)} to predict future trajectories from video and text, without any task-specific fine-tuning or trained prediction head. The large performance gap to ThinkJEPA indicates that zero-shot prompting is not sufficient for fine-grained metric-space trajectory forecasting. This study is included as a weak reference point only; the \emph{VLM-only} baseline in the main paper is a substantially fairer comparison because it is trained with the same task-specific supervision and downstream head.}
\label{tab:prompt_only}
\end{table}

\begin{table}[ht]
\centering
\small
\begin{tabular}{l l}
\hline
\textbf{Hyperparameter} & \textbf{Value} \\
\hline
Input frames ($T$) & 64 \\
Past/Future split ($T_p/T_f$) & 32/32 \\
Input resolution & $256{\times}256$ \\
Backbone & V-JEPA-L (\texttt{vit\_large\_rope}) \\
Backbone depth / dim & 24 / 1024 \\
Patch embedding & \texttt{Conv3d} kernel/stride $(2,16,16)$ \\
Predictor & VLM-injected V-JEPA predictor \\
Predictor dim ($D_p$) & 384 \\
Predictor depth / heads & 12 / 6 \\
RoPE / mask tokens & enabled / 2 \\
VLM thinker & \texttt{Qwen3-VL (Thinking)} (cached) \\
VLM token dim ($D_c$) & 2048 \\
Cache clips ($N_c$) & 8 \\
Encoder token length ($L_{\text{enc}}$) & 480 \\
AR token length ($L_{\text{ar}}$) & 15 \\
Pyramid layers ($\mathcal{L}$) & $\{0,4,8,12,16,20,24,27\}$ \\
Guidance injection & layer-wise FiLM \\
Temporal downsampling & AvgPool stride 2 (64$\rightarrow$32) \\
Output shape & $32 \times 52 \times 3$ \\
\hline
\end{tabular}
\caption{Key architectural hyperparameters and tensor dimensions.}
\label{tab:impl_hparams}
\end{table}

\section{Implementation Details}
\label{sec:supp_impl}

\subsection{Shared implementation setting.}
Tab.~\ref{tab:impl_hparams} summarizes the key architectural hyperparameters and tensor dimensions used throughout the supplementary experiments.
Unless otherwise specified, all experiments share the same base configuration: a 64-frame input clip at resolution $256\times256$, a V-JEPA-L backbone for latent token extraction, and a VLM-injected V-JEPA predictor operating in a latent dimension of $D_p=384$.
The VLM thinker is instantiated with cached \texttt{Qwen3-VL (Thinking)} features, including both encoder tokens and autoregressive tokens, and multi-depth VLM representations are extracted from the pyramid layer set $\mathcal{L}=\{0,4,8,12,16,20,24,27\}$.
Guidance is injected into the predictor via layer-wise FiLM modulation, and the final latent sequence is decoded through temporal downsampling to produce $32\times52\times3$ trajectory outputs.
This table is provided to clarify the common experimental backbone shared by the controlled ablations in the supplementary material.

\section{Limitations and Future Directions}
\label{sec:limitations}

ThinkJEPA marks a step toward unifying a cortex-like VLM for high-level semantic understanding with a cerebellum-like JEPA branch for low-level latent dynamics.
By connecting JEPA-style latent prediction with VLM-based semantic guidance, ThinkJEPA opens several directions that are not fully explored in this work.
First, while our experiments cover egocentric trajectory prediction, robotic latent rollout, and physical-scene forecasting, the current work focuses on predictive latent modeling rather than closed-loop planning or policy deployment.
Extending ThinkJEPA toward more agentic settings, where latent world models are used for planning, interaction, and decision-making in complex environments, is an important future direction.
Second, current latent world models such as V-JEPA2 and DINO-WM still face challenges in generalizing across diverse scenarios, and stronger VLM thinkers with broader world knowledge may further improve such generalization.
Finally, scaling ThinkJEPA to longer videos and more diverse embodied environments may require more efficient VLM feature caching and more adaptive temporal guidance mechanisms.
More broadly, we hope this work inspires further research on bridging semantic reasoning and latent world modeling with larger and more diverse embodied video datasets.

\section{Broader Impacts}
\label{sec:broader_impacts}

ThinkJEPA studies VLM-guided latent forecasting for video-based world modeling.
Potential positive impacts include improving video understanding, embodied AI, robot learning, and physical-scene forecasting systems that require compact predictive representations rather than expensive pixel-level generation.
A key motivation of this work is that existing latent world models may lack high-level semantic grounding, common-sense knowledge, and language-guided context when forecasting future states from video alone.
By introducing a VLM thinker as a source of broad visual--semantic guidance, ThinkJEPA may help reduce forecasting errors caused by purely local visual extrapolation, especially in settings that require long-horizon context, object-level semantics, or physical commonsense.

The main societal risks arise when predictive world models are deployed in downstream decision-making or robotics systems without sufficient validation.
Incorrect forecasts, biased data, or distribution shift may lead to unsafe or unfair decisions, particularly in human-centered environments.
Egocentric and embodied video data may also raise privacy concerns if collected or used without appropriate consent and safeguards.
Potential mitigations include careful dataset governance, privacy-preserving data handling, uncertainty estimation, human oversight, and restricting deployment in safety-critical settings until the forecasting model is validated under diverse real-world conditions.
Overall, we view VLM-guided latent forecasting as a step toward more semantically grounded and potentially safer world models, rather than as a standalone decision-making system.



\end{document}